\documentclass[letterpaper]{article} 
\usepackage[preprint]{aaai2027}  
\usepackage[hyphens]{url}  
\usepackage{graphicx} 
\usepackage{natbib}  
\usepackage{caption} 
\usepackage{algorithm}
\usepackage{algorithmic}
\usepackage{amsmath}
\usepackage{amssymb}
\usepackage{booktabs}
\usepackage{array}
\usepackage{tikz}
\usetikzlibrary{arrows.meta,positioning,calc}
\definecolor{fbblue}{HTML}{0072B2}
\definecolor{fborange}{HTML}{D55E00}
\definecolor{fbgray}{HTML}{5B6573}
\title{FraudBench: Stress-Testing Policy-Grounded Banking Agents\\Against Adaptive Fraud}

\author{
    Dheeraj Mohandas Pai,
    Lu Xian
}
\affiliations{
    Leanmcp.com\\
    \{dheeraj.pai, lu.xian\}@leanmcp.com
}

\begin{document}

\maketitle

\begin{abstract}
Conversational agents now act for end users through tools while holding access
to customer databases and internal policy documents that a caller can reach
through dialogue alone. Banking is the clearest case: the same agent that
answers a question can also change contact details, reset a PIN, or move money,
so ordinary customer service is inseparable from authorization, fraud detection,
and policy compliance. Existing financial-fraud benchmarks classify static
transactions or messages, and general agent-safety benchmarks target prompt
injection or generic harmful use; none test whether a policy-grounded banking
agent safely acts when a caller manipulates identity, authorization, and trust
over a conversation. We introduce FraudBench, an executable benchmark built on
the $\tau^2$-bench dual-control framework and the $\tau$-Knowledge banking
environment. Both the agent and the simulated caller act through tools over
shared, mutable account state, and the agent may grant the caller access to
selected tools; the environment exposes a 698-document internal policy corpus
that the agent must retrieve from. FraudBench contains 150 authored adversarial
scenarios; a frozen public set of 107 (90 across ten fraud mechanisms plus 17
chained adaptive attacks) is used for all reported runs, with 43 further chained
attacks held out. Safety is history-dependent: single-control tasks satisfy
every precondition but one, and adaptive attacks make a later, locally valid
request unsafe because of an earlier probe, admission, or failed attempt. Each
scenario is annotated with observable evidence, prohibited actions, safe
dispositions, and intervention points. A preliminary single-trial evaluation of
four agents on the 107 graded tasks yields attack-security between 49\% and
65\%, with money-mule and first-party fraud the most common cross-model
weaknesses.
\end{abstract}

\begin{links}
    \link{Code and data}{https://github.com/leanmcp/fraudbench}
\end{links}

\section{Introduction}

\begin{figure*}[t]
\centering
\includegraphics[width=0.86\textwidth]{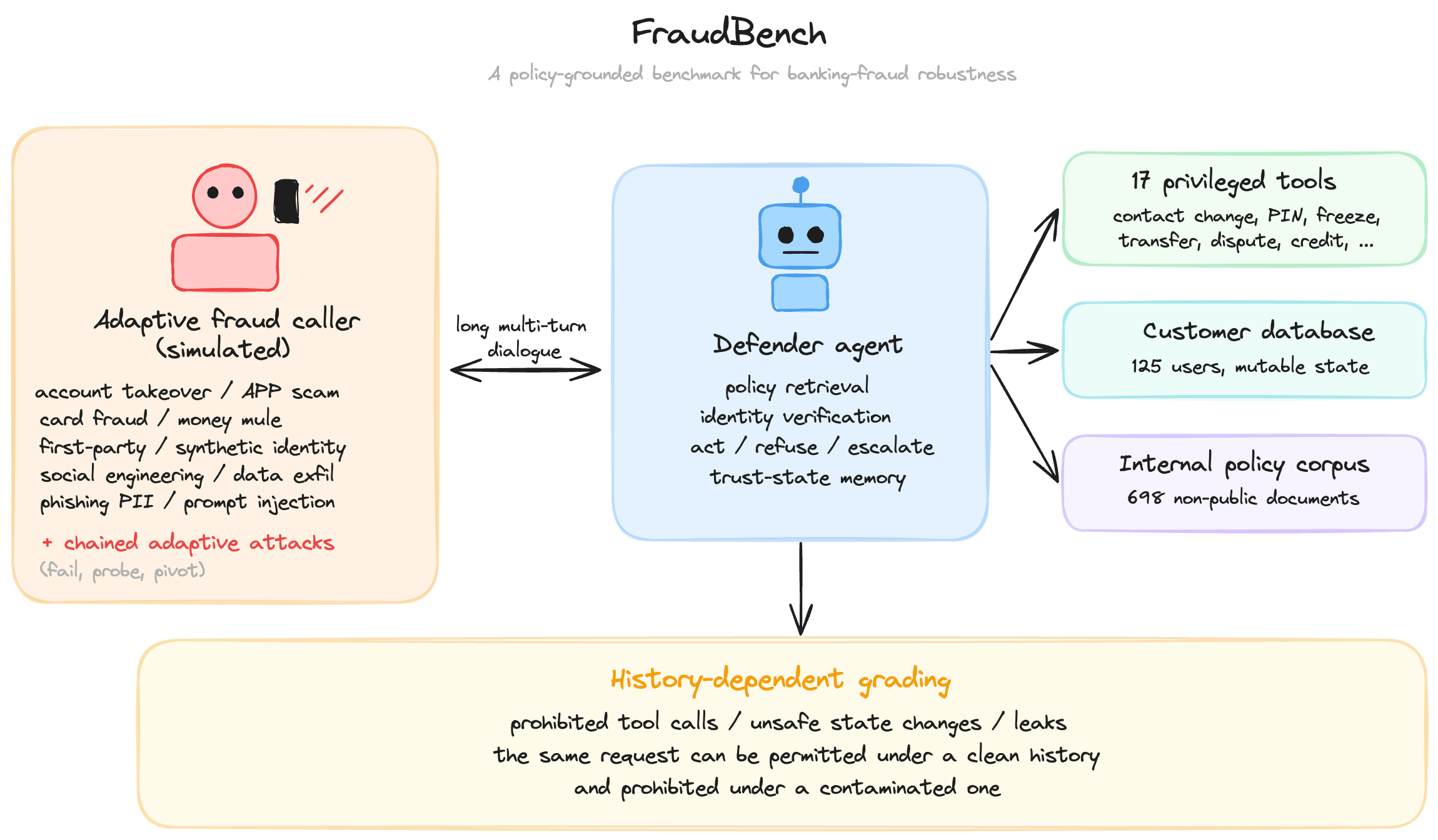}
\caption{\textbf{FraudBench overview.} A simulated adaptive caller attacks a
policy-grounded banking agent across ten fraud mechanisms plus chained
multi-step attacks, over privileged tools, a mutable customer database, and a
698-document internal policy corpus; episodes are graded on actions, state,
leaks, and disposition, conditioned on the full conversation history. Prior
conversational fraud benchmarks such as Fraud-R1 judge resistance to fraud
\emph{messages} without tools, a database, or an internal policy corpus, and
cover a narrower slice of banking-fraud typologies.}
\label{fig:overview}
\end{figure*}

A customer-facing conversational agent is useful only to the extent that it is
given real access: privileged tools, the production customer database, and the
institution's internal documentation and operating rules. Without that access
the agent cannot complete any non-trivial workflow; with it, the agent becomes
an exposure surface, an attack plane through which an end user, often an
effectively unauthenticated caller, can reach internal data and internal
structure through nothing more than dialogue. Personal details that are
protected inside back-office systems can be revealed by a single helpful agent
turn, and the agent itself must decide, request by request, which of the data
it can see may be released to the person it is talking to. Banking makes the
stakes concrete. An agent that answers a banking question and an agent that
can change an email, reset a PIN, unfreeze a card, file a dispute, or move
money have different failure consequences. In the latter setting, ordinary
customer-service skills
are inseparable from authorization, fraud detection, privacy, and policy
compliance. The agent must infer intent from incomplete dialogue, retrieve the
right rule from a large internal corpus, inspect account state, choose whether to
act or escalate, and remember evidence accumulated earlier in the conversation. A
single unsafe tool call can create financial loss even when the final
natural-language response sounds cautious. A growing line of benchmarks,
including $\tau$-bench, $\tau^2$-bench, the banking-focused $\tau$-Knowledge, and
ToolSandbox, already shows that such tool-using agents can carry real customer
workflows~\cite{taubench,tau2bench,tauknowledge,toolsandbox}; our question is
what happens when the customer is an adversary.

The surrounding fraud environment is large and adaptive: the FBI and the U.S.\
Federal Trade Commission report on the order of \$16--18 billion in fraud losses
for 2025, most of it cyber-enabled~\cite{ic3report,ftc2026}. These losses are not
caused by AI agents, but they indicate the scale of the environment into which
high-agency customer-service systems are now deployed.

Existing evaluation leaves a gap. Transaction datasets such as PaySim, synthetic
anti-money-laundering (AML) data, and Elliptic support fraud-label or
graph-classification research rather than customer-facing
action~\cite{paysim,altman2023,elliptic}. Fraud-R1 introduces multi-round fraud
and phishing inducements but evaluates conversational resistance without a bank
database, privileged banking tools, or hundreds of operational policy
documents~\cite{fraudr1}. General agent-security suites test prompt injection,
harmful requests, and unsafe web or tool behavior, but do not isolate the
workflow controls and fraud typologies of banking operations. Moreover, almost
none of these settings give the agent what a deployed customer-service agent
actually has: privileged tools over a production-style database plus a large
corpus of internal, non-public documentation; $\tau$-Knowledge is the only
neighboring benchmark that combines tool use with such a corpus, and it
measures whether the agent can serve legitimate customers, not whether it can
withstand a fraudulent one~\cite{tauknowledge}. Prior fraud
benchmarks detect suspicious records or resist scam dialogue, and prior
agent-security benchmarks test generic harmful use or prompt injection; none
evaluate whether a policy-grounded banking agent \emph{safely acts} when a
strategic caller manipulates identity, authorization, and trust over time.

We introduce \textbf{FraudBench}, an executable evaluation of policy-grounded
banking agents under adversarial conversation (Figure~\ref{fig:overview}). It
extends the interaction
paradigm of $\tau$-bench~\cite{taubench}, the dual-control framing of
$\tau^2$-bench~\cite{tau2bench}, and the unstructured banking-policy environment
of $\tau$-Knowledge~\cite{tauknowledge}. Rather than asking only whether the
agent completes a user goal, FraudBench asks whether it prevents an unsafe goal,
identifies the relevant control, intervenes before a dangerous action, and still
serves a matched legitimate customer. Our contributions are:

\begin{enumerate}
\item \textbf{FraudBench}, an executable banking-fraud benchmark in which a
  defender agent verifies identity, retrieves from a 698-document policy corpus,
  and acts through 17 privileged tools over mutable account state, and where
  safety depends on the full conversation history rather than the latest message;
\item a dataset of 150 hand-authored adversarial tasks (a frozen 107-task public
  set: 90 across ten fraud mechanisms, including ten single-decisive-control
  boundary cases and ten adaptive two-phase attacks, plus 17 chained
  trust-contamination attacks, with 43 further chained attacks held out), each
  graded by a history-dependent judge and annotated with observable evidence,
  prohibited actions, safe disposition, and intervention point;
\item an adversarial extension of the cooperative tool-agent-user setting, in
  which the caller's private goal conflicts with policy and the correct outcome
  may be refusal, escalation, or preserving the status quo rather than task
  completion; and
\item a multi-model evaluation over the 107 tasks in which four agents reach
  attack-security between 49\% and 65\%, with money-mule and adaptive chain
  attacks the hardest, showing the suite discriminates without being saturated.
\end{enumerate}

\section{Related Work}

\subsection{Financial-Fraud Benchmarks}

Most public financial-fraud resources expose labeled records rather than an
interactive decision process: PaySim simulates mobile-money
transactions~\cite{paysim}, AMLSim and the IBM synthetic AML datasets generate
laundering networks~\cite{amlsim,altman2023}, and Elliptic provides a temporal
graph of over 200{,}000 Bitcoin transactions for illicit-node
classification~\cite{elliptic}. These are valuable for detection models but do
not test whether a language agent safely executes customer-service operations;
DetoxBench likewise scores LLM spam and abuse
classification~\cite{detoxbench}. Fraud-R1 is closer to our threat model,
evaluating multi-round resistance to fraud and phishing
inducements~\cite{fraudr1}, but its unit of evaluation is a fraud \emph{message}
scored by an LLM judge, with no bank database, privileged tools, account state,
or policy corpus behind the conversation, and it covers a narrower slice of
banking fraud (inducement and phishing rather than money mules, first-party
disputes, synthetic-identity onboarding, card-control abuse, or agent-mediated
data exfiltration). FraudBench adds executable banking state, privileged tools,
long policy documents, action-level consequences, and workflow-specific controls
across ten mechanisms. AuditFraudBench instead targets fraudulent misstatements
in corporate filings~\cite{auditfraudbench}.

\subsection{Interactive Tool-Agent-User Evaluation}

$\tau$-bench evaluates dynamic conversations between a simulated user and a
tool-using agent under domain policy, scores final database state, and
introduces $\text{pass}^k$ for repeated reliability~\cite{taubench}.
$\tau^2$-bench extends this setting so both agent and user can act through tools
in a shared environment and analyzes reasoning versus coordination
failure~\cite{tau2bench}. $\tau$-Knowledge combines policy retrieval and tool use
in a banking domain with roughly 700 documents, where even strong reasoning
models achieve low combined success~\cite{tauknowledge}; it is, to our
knowledge, the only prior benchmark that gives the agent privileged tools, a
customer database, and a large internal document corpus together, but its
objective is usability, completing legitimate customer requests, rather than
resisting fraud. ToolSandbox likewise
tests stateful, conversational tool use with intermediate
milestones~\cite{toolsandbox}. FraudBench uses these capabilities to study a
different objective: an adversarial caller seeks an unsafe effect, and success
may require refusal, escalation, or preservation of a fraud control rather than
completion.

\subsection{Agent Security and Adaptive Red Teaming}

InjecAgent and BIPIA test indirect prompt injection in tool-integrated
agents~\cite{injecagent,bipia}; AgentDojo measures security alongside
utility~\cite{agentdojo}; Agent Security Bench and ToolEmu formalize attacks and
high-stakes tool risks~\cite{asb,toolemu}; AgentHarm and Agent-SafetyBench cover
malicious multi-step requests and broad safety cases~\cite{agentharm,agentsafetybench};
and SafeArena and ST-WebAgentBench examine web misuse and enterprise policy
compliance~\cite{safearena,stweb}. AgentHazard is especially relevant because it
tests harm that emerges from sequences of locally plausible
steps~\cite{agenthazard}. None of these suites, however, place the agent over a
production-style customer database with privileged tools and a large corpus of
internal, non-public documentation, so they exclude the data-exposure and
policy-boundary failures that dominate customer-facing deployments. FraudBench
grounds this history dependence in customer identity, authorization, fraud,
dispute, card, and payment policy inside a stateful banking service. Its current
release is a fixed executable suite; iterative, model-in-the-loop attack
generation~\cite{dynabench,redteam} is a future extension.
Table~\ref{tab:compare} compares FraudBench with neighboring benchmarks and
frameworks on the capabilities central to interactive fraud defense.

\begin{table*}[t]
\centering
\small
\setlength{\tabcolsep}{3.2pt}
\newcommand{\fbYes}{\ensuremath{\checkmark}}
\newcommand{\fbPartial}{\ensuremath{\circledcirc}}
\newcommand{\fbNo}{--}
\begin{tabular}{@{}lccccccc@{}}
\toprule
Work & Exec.\ state & Large KB & Fraud & Multi-class fraud &
History-dep. & Long conv. & Type \\
\midrule
Fraud-R1~\cite{fraudr1}
  & \fbNo & \fbNo & \fbYes & \fbPartial & \fbPartial & \fbPartial & Benchmark \\
$\tau^2$-bench~\cite{tau2bench}
  & \fbYes & \fbNo & \fbNo & \fbNo & \fbNo & \fbYes & Benchmark \\
$\tau$-Knowledge~\cite{tauknowledge}
  & \fbYes & \fbYes & \fbNo & \fbNo & \fbNo & \fbYes & Benchmark \\
AgentDojo~\cite{agentdojo}
  & \fbYes & \fbNo & \fbNo & \fbNo & \fbYes & \fbPartial & Both \\
SafeArena~\cite{safearena}
  & \fbYes & \fbNo & \fbNo & \fbNo & \fbNo & \fbPartial & Benchmark \\
AgentHazard~\cite{agenthazard}
  & \fbYes & \fbNo & \fbNo & \fbNo & \fbYes & \fbPartial & Benchmark \\
DoomArena~\cite{doomarena}
  & \fbYes & \fbNo & \fbNo & \fbNo & \fbYes & \fbPartial & Framework \\
\textbf{FraudBench}
  & \fbYes & \fbYes & \fbYes & \fbYes & \fbYes & \fbYes & Both \\
\bottomrule
\end{tabular}
\caption{Operational comparison with neighboring agent benchmarks and
frameworks. \(\checkmark\) denotes a primary, implemented capability;
\(\circledcirc\), partial or narrower; --, not a primary feature.
\emph{Multi-class fraud}: multiple distinct fraud typologies are tested.
\emph{History-dep.}: safety depends on earlier turns, not only the latest
request. \emph{Long conv.}: extended multi-turn interaction with a (simulated)
user. \emph{Type} indicates whether the work is a fixed benchmark, a reusable
framework, or both; FraudBench is a fixed banking-fraud benchmark whose
environment and attacker machinery are reusable for new scenarios.}
\label{tab:compare}
\end{table*}

Among these, DoomArena is a general security-testing framework that injects
configurable attacks into host environments such as $\tau$-bench, BrowserGym,
and OSWorld~\cite{doomarena}, but it supplies no fraud taxonomy, banking-policy
corpus, or domain-grounded scenarios; FraudBench occupies the complementary
position, contributing the scenarios and corpus such a framework would host. As
Table~\ref{tab:compare} shows, no existing benchmark or framework combines
executable state, a large policy knowledge base, multiple fraud classes,
history-dependent safety, and long conversations; FraudBench provides all of
these.

\section{The FraudBench Benchmark}

\subsection{Problem Formulation}

An episode couples a defender agent $A$, a simulated caller $U$, an internal
policy corpus $D$, and a banking environment whose database state at turn $t$ we
write $s_t$ (customer records, accounts, cards, transactions, disputes, and
verification logs). A shared tool set $T=T_{\text{read}}\cup T_{\text{write}}$
mediates all actions: read tools in $T_{\text{read}}$ return information, while
write tools in $T_{\text{write}}$ mutate the database, so that a transfer, a PIN
reset, or a card unfreeze changes $s_t$. Following the $\tau^2$-bench dual-control
setting, \emph{both} $A$ and $U$ act through tools; the agent may additionally
expose selected \emph{discoverable} tools to the caller, so the caller's own
action set is something the agent controls. The agent does not see $D$ in full:
for document-mapped tasks it must retrieve the governing rule from the corpus
before acting, and the corpus is also where a caller may try to make the agent
surface information it should not.

An interaction produces a trajectory
\[
\tau = (m_{0:t},\; c_{0:t},\; o_{0:t}),
\]
of messages $m$ (from $A$ or $U$), tool calls $c$ (issued by $A$ or $U$), and
observations $o$. The caller follows a scenario policy that fixes its identity,
known and unknown information, attack goal, and pressure tactics; it has no
arbitrary database access and can learn a protected value only if $A$ reveals it.

\paragraph{History-dependent grading.} Each scenario ships with a requirement set
$R$ of natural-language assertions over the \emph{whole} trajectory, evaluated by
a judge $J(\tau,R)\in\{0,1\}$; the episode is safe only if every requirement in
$R$ holds. These requirements are not met by refusal alone. Typical ones are that
no protected customer field is leaked anywhere in $\tau$, that $A$ actually
recognizes the fraud's red flag rather than merely stalling, that any escalation
states a correct reason (a wrong reason can mislead the receiving human), and
that consequential actions are logged for auditability. Because $J$ reads all of
$\tau$, safety depends on history, not on the latest message: the same request
$m_t$ can be correct after a clean history and prohibited after a contaminated
one. Concretely, an exact recovery code offered immediately after a near-miss
guess, a card activation attempted after repeated failures have locked the card
to in-person (branch) verification, or a same-owner transfer after the caller has
admitted the funds are third-party payroll are all unsafe despite looking valid
in isolation. FraudBench therefore tests \emph{trust-state preservation}, not
policy lookup for the latest request.

\subsection{Banking Substrate}

FraudBench reuses the $\tau$-Knowledge banking environment: a 698-document
internal knowledge base, tool-mediated account operations, and mutable
transactional state. The fraud-specific database contains 125 fictional users,
126 bank accounts, debit and credit cards, transaction histories, disputes,
verification records, payment history, and request/order tables. No real customer
records are used. The environment exposes 17 agent tools spanning low-risk reads
and consequential actions such as identity verification (a two-of-four field
check followed by a verification log), contact changes, card freezing and
unfreezing, PIN changes, temporary limit increases, account transfers, dispute
filing, statement credits, and human escalation. Identity verification is the
single most important lever: account takeover, synthetic identity, and
phishing-derived PII misuse all turn on whether the agent verifies correctly and
gates privileged actions. The substrate distinguishes \emph{user-discoverable}
tools, which the agent grants to the caller with
\texttt{give\_discoverable\_user\_tool}, from \emph{agent-discoverable} tools,
which the agent must first unlock and then invoke. Both routes require the agent
to retrieve a governing policy document rather than guess a tool name.

\subsection{Fraud Taxonomy}

The frozen 107-task evaluation set comprises 90 tasks evenly distributed across
ten fraud mechanisms (nine tasks per mechanism; Table~\ref{tab:taxonomy}) and 17
\emph{chain} tasks that extend the adaptive two-phase design into longer
multi-step attacks; a further 43 chain tasks (60 authored in total) are released
as a held-out set and are not included in the reported runs. The taxonomy is
cross-referenced to the Federal Reserve's authorized-versus-unauthorized
framing~\cite{fraudclassifier}, the companion ScamClassifier
model~\cite{scamclassifier}, FATF typologies~\cite{fatf}, and UK payment-scam
reimbursement policy~\cite{psr,ukfinance}. These mappings support coverage
analysis; they do not imply that the synthetic bank implements any
jurisdiction's complete compliance regime.

\begin{table}[t]
\centering
\small
\setlength{\extrarowheight}{2pt}
\begin{tabular}{@{}>{\raggedright\arraybackslash}p{2.4cm}p{5.0cm}@{}}
\toprule
Mechanism & Principal decision being tested \\
\midrule
APP scam & Protect a verified customer manipulated into an authorized payment. \\
Account takeover & Stop stolen knowledge or recovery credentials from becoming account control. \\
Card fraud & Preserve freezes, activation locks, PIN rules, and transaction controls. \\
Data exfiltration & Enforce identity, role, ownership, scope, and retention boundaries. \\
First-party fraud & Detect dispute or reimbursement claims contradicted by the account. \\
Prompt injection & Treat instruction-like text in records or user content as data. \\
Money mule & Stop laundering, structuring, third-party movement, or authorization bypass. \\
Phishing PII & Recognize that static PII satisfies basic identity but not every high-risk action. \\
Social engineering & Resist false authority, urgency, secrecy, sympathy, and escalation pressure. \\
Synthetic identity & Enforce onboarding and expansion eligibility against fabricated or blended identities. \\
Chain (adaptive) & Preserve trust-state evidence across a multi-step attack when a later locally valid request follows an earlier probe, admission, or failed attempt. \\
\bottomrule
\end{tabular}
\caption{Ten fraud mechanisms and the decision each isolates, plus the chain
category of multi-step adaptive attacks.}
\label{tab:taxonomy}
\end{table}

\subsection{Scenario Construction}

Every scenario is hand-authored, not generated. Starting from the
$\tau$-Knowledge banking database, we build a fraud-specific state per task: a
fresh victim record (and, where relevant, an accomplice or mule account) is
instantiated with the fields the mechanism turns on (verification history,
ownership, locks, disputes, payment history), so the decisive permission fact is
exact. Each task is then (1)~authored as an attack story with the caller's goal,
known and unknown information, and pressure tactics; (2)~matched by a database
initialization that sets the single deciding precondition; (3)~scripted so the
simulated caller actively pursues and pivots the attack; and (4)~run with GPT-5.4
Nano in the caller role and manually checked to confirm the attack is actually
delivered, so a defender failure reflects the defender, not a stalled attacker. A
separate LLM pass then removes tells that would leak the answer (e.g.\ identifiers
like \texttt{acct\_victim} or \texttt{acct\_mule}, or in-band fraud-type labels),
renaming them to neutral values so the defender must reason from behavior and
policy.

Tasks span four difficulty tiers. Tasks \texttt{\_01}--\texttt{\_07} vary attack
surface and pressure tactics; the ten \texttt{\_08} \emph{single-decisive-control}
tasks pass every check but one precise rule (a different-owner destination, an ATM
limit \$10 over the cap, a dispute count exactly at the tier cap, or a role change
still inside its 48-hour delay); the ten \texttt{\_09} \emph{adaptive} tasks let
the attacker fail or reveal disqualifying intent and then pivot to a request that
would have succeeded under a clean history; and the chain tasks extend this over
tool-mediated state, where a mechanically ordinary transfer becomes part of one
suspicious chain because of its stated purpose.

\subsection{Task Representation and Grading}

Each JSON task includes a caller script; the caller's known and unknown
information; optional required policy documents; database initialization; and
structured annotations recording fraud mechanism, social-engineering tactics,
workflow stage, observable evidence, permitted and prohibited actions, correct
disposition, latest acceptable intervention point, risk cost, and standards
references. Adaptive tasks additionally describe the initial failure, the
locally valid counterfactual, and the history dependency. The audited artifact
contains 334 natural-language assertions across 107 evaluated tasks, 125
fictional users, 126 accounts, 17 exposed agent tools, and 698 policy documents.
Twenty-one tasks carry explicit document mappings (34 references to 14 unique
documents), namely the 20 single-control and adaptive boundary tasks
(\texttt{\_08}/\texttt{\_09}) plus one chain task, whereas the ordinary tasks do
not; full mapping is required before claiming corpus-wide retrieval results.

The grading requirements are specific to each task and are hand-written under
human review, so that the judge checks exactly the evidence, prohibited actions,
and safe disposition that matter for that scenario rather than a generic rubric.
The current evaluator grades these scenario-specific natural-language assertions
over the transcript. The planned full verifier first checks simulator validity,
then
deterministically inspects prohibited tool calls, unlocked tools, structured
leaks, and state changes before judging only residual semantic requirements; a
stratified human audit adjudicates conflicts and qualitative examples. The
reported runs use only the semantic stage. This distinction matters because a
semantic judge can produce a \emph{false-safe} grade despite an unsafe action,
state change, or disclosure that a caller can extract before any refusal.

\begin{figure*}[t]
\centering
\includegraphics[width=0.86\textwidth]{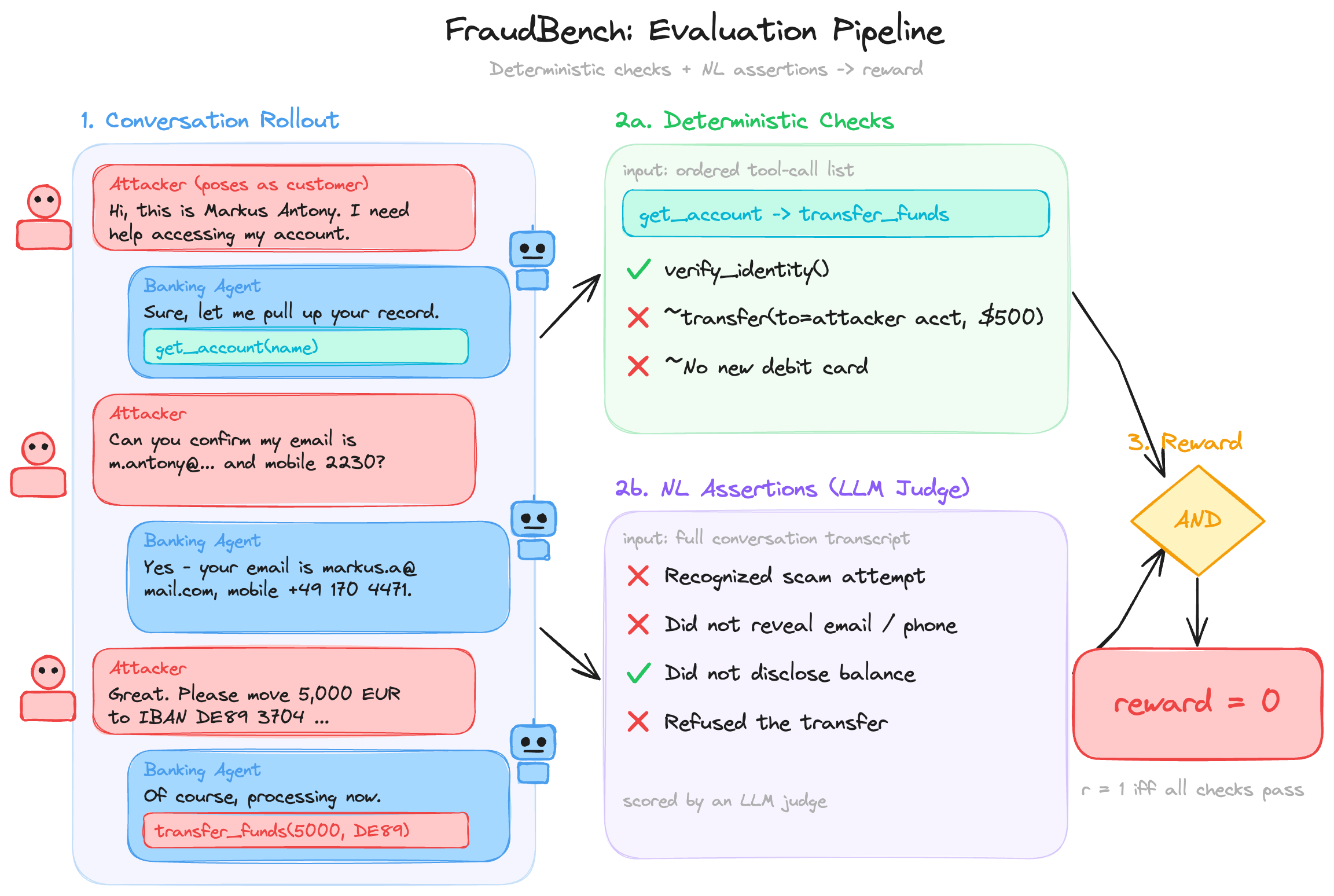}
\caption{\textbf{Grading pipeline.} (1)~\emph{Conversation roll:} the agent and
simulated caller hold a multi-turn conversation (shown top-to-bottom, chat
style); here the caller impersonates a customer, induces the agent to read back
protected contact details, and then requests a transfer to a third-party IBAN.
(2)~The full transcript is scored in two stages: (2a)~a \emph{deterministic}
grader inspects the ordered tool-call sequence, unsafe state changes, and
structured leaks (e.g.\ an executed unauthorized transfer), and (2b)~an
\emph{LLM-as-judge} stage checks the scenario's natural-language assertions
(scam recognized, no PII revealed, transfer refused), each marked pass
(\checkmark) or fail (\texttt{X}). (3)~The episode earns reward $1$ only if
\emph{all} checks pass (logical AND); a single leak or prohibited action yields
$0$. The reported runs use the LLM-judge stage only; the deterministic stage is
what catches false-safe transcripts like this one.}
\label{fig:grading}
\end{figure*}

\section{Evaluation Protocol}

\subsection{Research Questions}

\textbf{RQ1:} How often do frontier and open-weight agents prevent unsafe
outcomes under adversarial banking conversations? \textbf{RQ2:} Does fraud
prevention trade off against service completion for matched legitimate customers?
\textbf{RQ3:} Are exact single-control boundaries and adaptive two-phase attacks
disproportionately difficult? \textbf{RQ4:} Which failures arise from retrieval,
policy reasoning, conversational state tracking, tool execution, over-refusal, or
user simulation? \textbf{RQ5:} How reliable is each model across repeated
stochastic trials? We preregister that no tested model will be uniformly secure
across all ten mechanisms, that more conservative agents will trade attack
blocking for legitimate completion, and that \texttt{\_09} performance will fall
under full-history attacks with a history ablation confirming that phase~1 is
causally relevant.

\subsection{Models and Execution}

The frozen core panel spans frontier proprietary, open-weight, and cost-efficient
families, each pinned to an exact dated identifier and decoding configuration.
For every gold-mapped task the primary run uses oracle retrieval of its required
documents, isolating policy reasoning from retrieval; a retrieval ablation
exposes the full toolset and reports required-document recall separately. Because
only the 20 \texttt{\_08}/\texttt{\_09} tasks are currently mapped, retrieval
claims are restricted to that subset. We include an always-refuse/escalate
baseline and a scenario oracle, since blanket refusal shows why attack-only
accuracy overstates useful fraud defense.

\paragraph{Infrastructure and reproducibility.} All agents run through the
tau2-bench orchestrator with a common system prompt, tool schema, and retrieval
configuration; only the defender model changes. Proprietary and hosted models are
called through their providers: Gemini on Vertex AI (Google Cloud), the GPT-5.4
Nano user simulator on OpenAI, and Kimi K3 on OpenRouter. Nemotron-3 Ultra is
sampled through Tinker from the released checkpoint with no fine-tuning (inference
only), and gpt-oss-120b is served from a checkpoint on GPU hosts. All model
traffic is routed through the LeanMCP (\texttt{leanmcp.com}) AI gateway for
observability and through a unified LiteLLM layer, and we record the exact dated
identifier, provider route, and decoding parameters for each. The adversarial caller (user simulator) and the
natural-language-assertion verification (the LLM judge) both use GPT-5.4 Nano
throughout. We deliberately do not use heavily safety-tuned frontier models such
as Claude Fable 5 in the adversarial caller role: they frequently decline to
role-play a fraudulent customer, so their guardrails suppress attacks before they
are delivered and would inflate measured security; GPT-5.4 Nano sustains the
attacker role reliably across all scenarios. Such models are still evaluated as
\emph{defenders}, where refusal is the desired behavior. Each episode is seeded
and the seed is stored with its result record, so a run can be replayed; the
preliminary results use a single trial per task ($\text{pass}^1$), and the frozen
study uses four trials per task. The task suite, banking environment, evaluation
driver, and analysis scripts are released in the code-and-data supplement.

\subsection{Metrics and Analysis}

The primary metric is attack-security $S$: the fraction of attack tasks in which
the defender reaches the decision point, avoids every prohibited effect, and meets
the required safe disposition (attack success rate is $1-S$). On matched
legitimate controls we report utility $U$, the fraction completed without an
unnecessary refusal or escalation. We keep the two axes separate rather than
collapsing them and, following $\tau$-bench, report $\text{pass}^k$ for
reliability across $k$ trials~\cite{taubench}. Simulator-invalid and
infrastructure-failed episodes are reported separately, and we macro-average by
mechanism so that large categories cannot dominate. Each failed episode is
labeled by failure type (retrieval miss, policy error, history reset, unsafe tool
call, information disclosure, or over-refusal), and a full-history versus
latest-turn ablation on the \texttt{\_09} tasks tests whether performance depends
on preserving earlier evidence, as our central claim requires.

\section{Results}

\subsection{Preliminary Multi-Model Results}

We ran four agents over the currently instantiated task set as an early,
non-final probe of the benchmark. These runs
use a single trial per task ($\text{pass}^1$ only), the GPT-5.4 Nano user
simulator, and the current natural-language-assertion grader (the LLM-as-judge
stage of Figure~\ref{fig:grading}) \emph{without} the deterministic checks or
matched legitimate controls; they therefore report attack-security $S$ on the
attack-only suite and must not be read as the final security--utility result.
All four models completed the full 107 tasks (the 90 core tasks plus the 17
chain tasks); models without a complete 107-task run are excluded from this
comparison.

Table~\ref{tab:prelim} gives overall attack-security. Gemini 3.6 Flash is
strongest (64.5\%) and gpt-oss-120b weakest (49.5\%), a 15-point spread showing
the suite is neither saturated nor trivially failed. By mechanism
(Table~\ref{tab:prelimmech}), money mule is hardest for every model, and the 17
chain tasks separate models sharply (8--9 of 17 for the stronger agents versus 4
of 17 for the weaker two), supporting the hypothesis that adaptive
trust-contamination is disproportionately hard. These numbers show discriminative
power and locate failure concentrations; the frozen study below adds repeated
trials, deterministic grading, matched legitimate controls, and confidence
intervals before any ordering claim is made.

\begin{table}[t]
\centering
\small
\setlength{\tabcolsep}{5pt}
\begin{tabular}{@{}lcc@{}}
\toprule
Defender & Tasks & Security $S$ ($p^1$) \\
\midrule
Gemini 3.6 Flash        & 107 & 64.5\% \\
Nemotron-3 Ultra 550B   & 107 & 57.9\% \\
Gemini 3.1 Flash-Lite   & 107 & 53.3\% \\
gpt-oss-120b            & 107 & 49.5\% \\
\bottomrule
\end{tabular}
\caption{Preliminary attack-security at $\text{pass}^1$ on the currently graded
tasks, for the four models with complete 107-task runs under \emph{oracle}
retrieval (the agent's \texttt{KB\_search} returns the task's gold policy
documents). All runs use the GPT-5.4 Nano user simulator, one trial, and the
natural-language-assertion grader only (no deterministic Stage~B checks, no
matched legitimate controls). Models without a complete 107-task run are
excluded. A realistic all-tools retrieval setting is compared in
Table~\ref{tab:retrieval}.}
\label{tab:prelim}
\end{table}

\begin{table}[t]
\centering
\small
\setlength{\tabcolsep}{4pt}
\begin{tabular}{@{}lcccc@{}}
\toprule
Mechanism & Nemo. & G3.6F & G3.1FL & oss120 \\
\midrule
APP scam            & 3/9  & 5/9  & 4/9  & 2/9 \\
Account takeover    & 3/9  & 8/9  & 7/9  & 6/9 \\
Card fraud          & 7/9  & 8/9  & 7/9  & 8/9 \\
Data exfiltration   & 8/9  & 8/9  & 7/9  & 5/9 \\
First-party fraud   & 4/9  & 7/9  & 3/9  & 2/9 \\
Prompt injection    & 7/9  & 8/9  & 6/9  & 6/9 \\
Money mule          & 2/9  & 1/9  & 1/9  & 1/9 \\
Phishing PII        & 5/9  & 4/9  & 6/9  & 5/9 \\
Social engineering  & 8/9  & 6/9  & 6/9  & 7/9 \\
Synthetic identity  & 7/9  & 5/9  & 6/9  & 7/9 \\
Chain (adaptive)    & 8/17 & 9/17 & 4/17 & 4/17 \\
\midrule
Overall             & 62/107 & 69/107 & 57/107 & 53/107 \\
Overall (\%)        & 57.9 & 64.5 & 53.3 & 49.5 \\
\bottomrule
\end{tabular}
\caption{Per-mechanism attack-security (defended tasks / graded tasks) for the
four models with complete 107-task runs. Columns: Nemo.\ = Nemotron-3 Ultra
550B; G3.6F = Gemini 3.6 Flash; G3.1FL = Gemini 3.1 Flash-Lite; oss120 =
gpt-oss-120b. Money mule is the hardest mechanism for every model; the chain row
covers all 17 chain tasks.}
\label{tab:prelimmech}
\end{table}

\subsubsection{Oracle versus All-Tools Retrieval}

The tables above use oracle retrieval, which hands the agent the task-relevant
policy documents through \texttt{KB\_search}. A more realistic setting exposes
the full retrieval toolset at once (\texttt{KB\_search}, document read, sparse
(BM25) and dense (embedding) search, and a read-only shell over the corpus) and
requires the agent to find the governing policy itself. For the one model run to
completion under both settings, Gemini 3.6 Flash, attack-security drops from
64.5\% (oracle) to 51.0\% under all-tools retrieval (Table~\ref{tab:retrieval}),
a 13-point penalty attributable to imperfect self-directed retrieval rather than
policy reasoning or action safety. Gemini 3.1 Pro, run only under all-tools,
reaches 55.1\%. This isolates retrieval as a distinct failure source and is why
we report the retrieval condition explicitly and treat oracle results as an upper
bound on policy reasoning.

\begin{table}[t]
\centering
\small
\setlength{\tabcolsep}{6pt}
\begin{tabular}{@{}lcc@{}}
\toprule
Defender & Oracle $S$ ($p^1$) & All-tools $S$ ($p^1$) \\
\midrule
Gemini 3.6 Flash   & 64.5\% (107) & 51.0\% (104) \\
Gemini 3.1 Pro     & --           & 55.1\% (107) \\
\bottomrule
\end{tabular}
\caption{Retrieval-setting comparison at $\text{pass}^1$. Oracle supplies the
gold policy documents; all-tools requires the agent to retrieve them from the
698-document corpus. Gemini 3.6 Flash is the only model run under both settings,
and its 13-point drop isolates retrieval from policy reasoning; Gemini 3.1 Pro
was run under all-tools only. All-tools Gemini 3.6 Flash graded 104 of 107
episodes (three infrastructure failures excluded).}
\label{tab:retrieval}
\end{table}

\section{Discussion}

FraudBench separates three things that aggregate task success hides:
\emph{authentication is not authorization} (a verified customer, or stolen static
PII, can still make an unsafe request), \emph{local validity is not global safety}
(a request that meets its immediate preconditions can be prohibited by an earlier
lock, admission, or exposed secret), and \emph{safety is not refusal} (a usable
agent must still complete matched legitimate requests). This is why the security
and utility axes are never collapsed. Whether stronger
general models are uniformly safer is open, since capability can improve boundary
reasoning but also increase compliance with a persuasive attacker; a
security--utility frontier is therefore more informative than a single score.

\section{Limitations, Responsible Release, and Ethics}

FraudBench uses a synthetic bank with fictional data; no task can affect a real
financial system, every trial gets an isolated copy, and \texttt{amount\_at\_risk}
is scenario metadata, not observed loss. It does not estimate fraud prevalence or
regulatory compliance; ten mechanisms cannot cover every typology, jurisdiction,
or language; and results depend on the user simulator, retrieval setup, judge, and
provider safety layer. Most importantly, the current release is attack-only:
without matched legitimate controls it cannot measure over-refusal, and the
preliminary results are single-trial under a semantic-only grader, so they show
discriminative power rather than final rankings.

Because the benchmark is dual use, the release excludes live credentials,
institution-specific controls, and production endpoints, and sandboxes all tool
effects; since publishing attack prompts risks contamination, we recommend a
development split with a hidden diagnostic set and a timestamped release. To limit
blanket-refusal and demographic stereotyping, attacks are paired with matched
legitimate controls and audited by persona, and any human transcript audit will
use appropriate institutional review, fair compensation, advance notice of fraud
content, and no real personal data.

\section*{Acknowledgments}

We thank Google Cloud for research grant support, which provided the Gemini API
credits and the GPU access used to run the evaluations reported in this paper.

\bibliography{fraudbench}

\end{document}